\documentclass{article}
\usepackage{ijcai23}

\usepackage{float}
\usepackage{times}
\usepackage{graphicx}
\usepackage{soul}
\usepackage{url}
\usepackage[hidelinks]{hyperref}
\usepackage[utf8]{inputenc}
\usepackage[small]{caption}
\usepackage{graphicx}
\usepackage{amsmath}
\usepackage{amsthm}
\usepackage{booktabs}
\usepackage{algorithm}
\usepackage{algorithmic}
\usepackage[switch]{lineno}
\usepackage{subfigure}
\usepackage{subcaption}
\usepackage{graphicx}
\usepackage{amssymb}

\title{PA-CLIP: Enhancing Zero-Shot Anomaly Detection through Pseudo-Anomaly Awareness}

\author{
Yurui Pan$^1$
\and
Lidong Wang$^1$\and
Yuchao Chen$^2$\and
Wenbing Zhu$^2$\and
Bo Peng$^*$\And
Mingmin Chi$^*$\
\affiliations
$^1$Fudan University\\
$^2$National University of Singapore\\
$^3$Shanghai Ocean University\\
\emails
\{24210240042, 23212010027\}@m.fudan.edu.cn,
chen.yuchao@u.nus.edu,
louis.zhu@rongcheer.com,
bepng@shou.edu.cn,
mmchi@fudan.edu.cn,
}

\begin{document}

\maketitle
\begin{abstract}
In industrial anomaly detection (IAD), accurately identifying defects amidst diverse anomalies and under varying imaging conditions remains a significant challenge. 
Traditional approaches often struggle with high false-positive rates, frequently misclassifying normal shadows and surface deformations as defects, an issue that becomes particularly pronounced in products with complex and intricate surface features.
To address these challenges, we introduce PA-CLIP, a zero-shot anomaly detection method that reduces background noise and enhances defect detection through a pseudo-anomaly-based framework. The proposed method integrates a multiscale feature aggregation strategy for capturing detailed global and local information, two memory banks for distinguishing background information, including normal patterns and pseudo-anomalies, from true anomaly features, and a decision-making module designed to minimize false positives caused by environmental variations while maintaining high defect sensitivity.
Demonstrated on the MVTec AD and VisA datasets, PA-CLIP outperforms existing zero-shot methods, providing a robust solution for industrial defect detection.
\end{abstract}

\section{Introduction}
Industrial anomaly detection (IAD) through defect detection plays a vital role in manufacturing processes. However, the absence of labeled training data, combined with the wide variety of potential product anomalies, presents major challenges for automated inspection systems in industrial settings.
In the field of anomaly detection, detection methodologies have undergone an evolution from conventional approaches to more sophisticated zero-shot techniques. Traditional methods include unsupervised approaches like PaDiM \cite{defard2021padim} and PatchCore \cite{roth2022towards}, which utilize feature reconstruction and memory bank to model normal patterns, and semi-supervised methods that leverage both normal and labeled anomalous samples to establish decision boundaries. To overcome the data dependency limitations of these methods, zero-shot anomaly detection (ZSAD) emerged, leveraging pre-trained Vision-Language Models (VLMs) like CLIP \cite{radford2021learning}. Recent works, such as APRIL-GAN and CLIP-AD \cite{chen2024clip}, have enhanced ZSAD performance through auxiliary dataset fine-tuning, while AnomalyCLIP \cite{zhou2023anomalyclip} and AdaCLIP \cite{cao2025adaclip} introduced optimized text prompts for better adaptation.

\begin{figure}[t] 
    \centering
    \includegraphics[width=\linewidth]{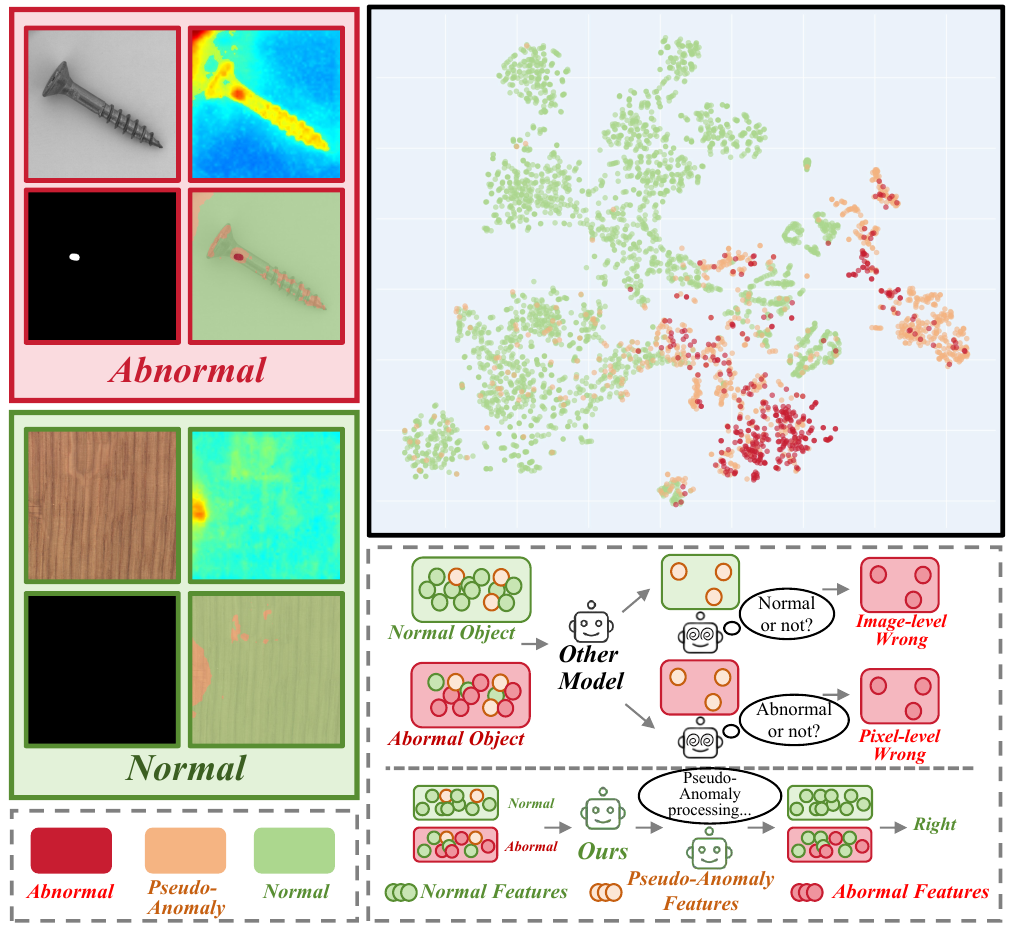} 
    \caption{Left: Example images from two categories are shown: anomalous images and entirely normal images. Green indicates normal regions, orange indicates pseudo-anomalous regions, and red indicates anomalous regions. The heatmap is generated by a SOTA model. Top Right: This subfigure illustrates the feature distribution after dimensionality reduction. Green points represent normal features, orange points represent pseudo-anomalous features, and red points represent true anomalous features. Bottom Left: Our model outperforms others in distinguishing pseudo-anomalies.
    }
    \label{fig:example}
\end{figure}

Despite the significant progress in anomaly detection methods, a critical challenge persists in handling environmental variations and noise. In particular, contrast-based methods frequently misclassify normal shadows and deformations, caused by varying shooting angles and lighting conditions, as product defects, leading to unacceptably high false-positive rates, especially in categories with complex surface features.
Based on our comprehensive statistical analysis of ViT-extracted features \cite{dosovitskiy2020image}, we observe distinct distribution patterns in the patch feature space: normal regions cluster tightly, true defects appear as outliers, while environmental variations exhibit regular offset patterns from the normal distribution. We define these environmentally-induced variations as pseudo-anomalies-variations that deviate from normal patterns due to environmental factors rather than actual defects. These pseudo-anomalies maintain consistent patterns under similar conditions and are typically reversible, fundamentally differentiating them from true structural defects. This key observation not only provides insights into the nature of different types of variations in industrial inspection scenarios but also drives our more robust approach to anomaly detection, effectively addressing the limitations of existing methods in handling complex real-world scenarios.

Drawing on these insights, we develop a pseudo-anomaly-aware memory bank that capitalizes on the characteristic distributions of pseudo-anomalies in the feature space. This approach enables accurate identification and subsequent removal of background noise through differential processing, substantially enhancing detection accuracy. Building upon this framework, we propose PA-CLIP, a zero-shot industrial defect detection method that operates solely with class names as prompts, eliminating the need for normal sample images. Our model architecture comprises three key components: a Multi-Scale Feature Aggregation module, a Pseudo-Anomaly-Aware Memory module, and a Pseudo-Anomaly-Aware Decision Module.
First, the Multi-Scale Feature Aggregation module employs dedicated encoders to process visual and textual inputs independently. By extracting visual features at multiple scales (r = 1, 3, 5) and computing their similarities with text embeddings, this module synthesizes both fine-grained details and holistic contextual information. This hierarchical feature aggregation strategy ensures comprehensive representation learning across different spatial resolutions.
Second, the Pseudo-Anomaly-Aware Memory module implements a dual memory bank architecture, comprising a background-specific and a comprehensive memory bank. Through systematic data augmentation and strategic core set sampling, this module facilitates robust discrimination between background characteristics and anomalous patterns. This architectural design enables effective feature disentanglement, enhancing the system's capability to isolate genuine defects from environmental variations.
Finally, the Pseudo-Anomaly-Aware Decision Module implements an adaptive fusion mechanism to integrate information from both memory banks. By computing the differential response between background-induced and comprehensive anomaly maps, this module effectively attenuates environmentally-induced artifacts (arising from illumination variations and viewpoint transformations) while preserving sensitivity to genuine structural defects. This differential analysis framework significantly enhances the robustness of anomaly detection in complex industrial environments.
Experimental results demonstrate that our method achieves significant performance improvements in industrial anomaly detection tasks, particularly in scenarios with complex background variations and subtle defect patterns. Compared to state-of-the-art zero-shot AC/AS methods \cite{li2024musc}, we achieve a +0.6\% AUROC classification metric on MVTec AD and a +0.8\% improvement on VisA. For anomaly segmentation, we obtain a +0.2\% PRO and +1.1\% AP gain on MVTec AD, and a +1.5\% AP gain on VisA. Moreover, our method virtually eliminates all background noise.

\begin{itemize}
 
\item To the best of our knowledge, we propose the first method for industrial defect detection that constructs a memory bank using only unlabeled test images.
 
\item We reveal the distribution relationship of background noise in the feature space and propose a Pseudo-Anomaly-Aware multi-scale feature extraction and fusion framework, PA-CLIP\@. This framework introduces Pseudo-Anomaly awareness for the first time in zero-shot industrial anomaly detection. Through multi-scale feature aggregation and dual memory bank design, it effectively captures visual information at different scales, achieving precise localization and segmentation of product defects.
 
\item Our method achieves optimal performance on both the MVTec AD and VisA datasets, reaching performance metrics of 98.4\% AUROC and 97.5\% pixel-AUROC on MVTec AD, and 93.4\% AUROC and 98.8\% pixel-AUROC on VisA datasets respectively, surpassing existing zero-shot methods. It demonstrates significant advantages, particularly in handling complex backgrounds and subtle defects.
 
\end{itemize}

\begin{figure*}[t]
    \centering
    \includegraphics[width=1.0\linewidth]{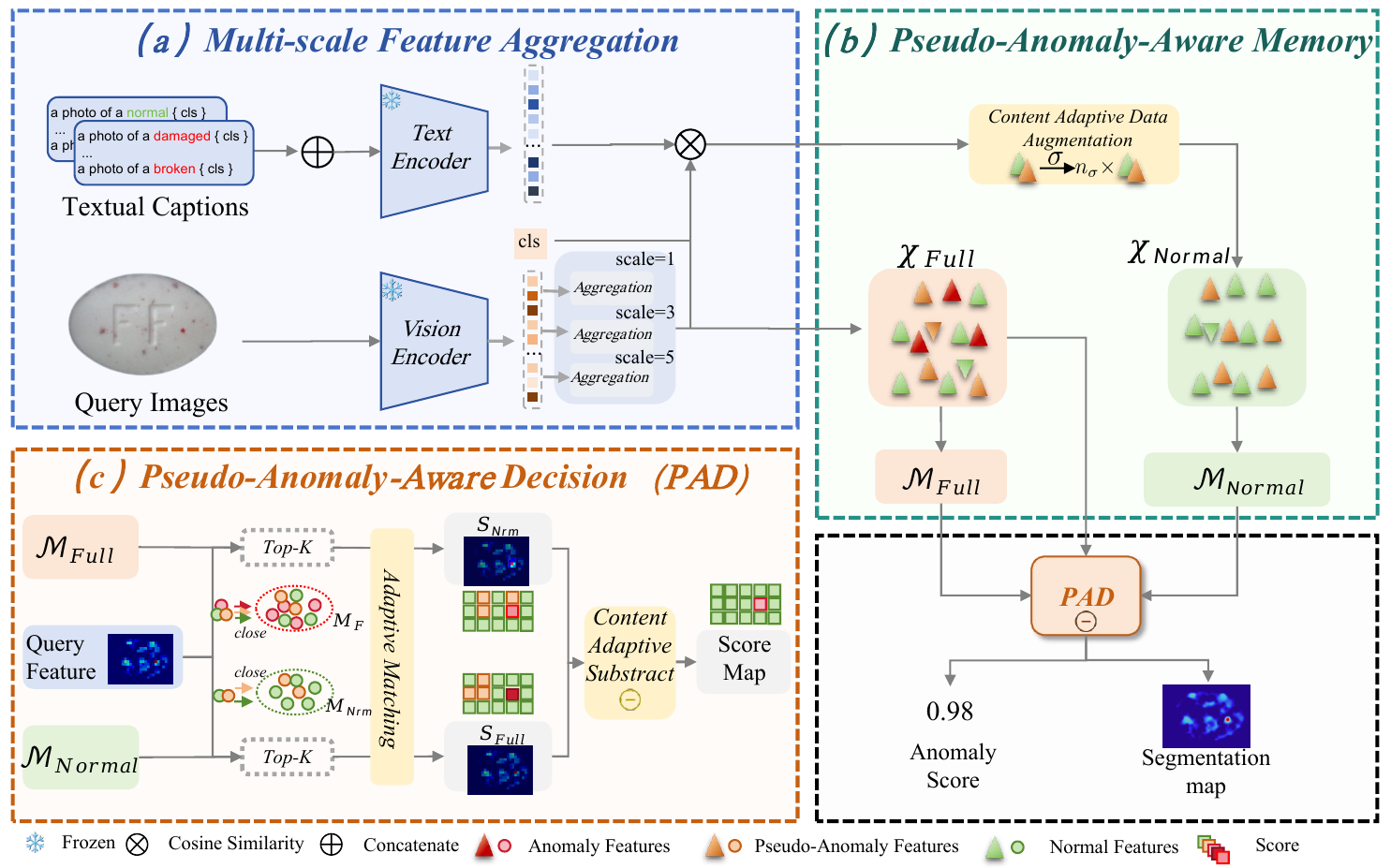}
    \caption{%
        \textbf{Overview of PA-CLIP.}
        It involves a three-stage process:
        (a)~Multi-Scale Aggregation,
        (b)~Pseudo-Anomaly-Aware Memory,
        (c)~Pseudo-Anomaly-Aware Decision.%
        }
    \label{fig:pipeline}
\end{figure*}
\section{Related Work}

\subsection{Anomaly Detection}
Anomaly Detection (AD) has been a critical focus in computer vision research, with methods generally categorized into unsupervised and semi-supervised approaches. Unsupervised AD techniques primarily rely on normal samples during the training phase, aiming to model the distribution of normal data and detect anomalies by identifying deviations from this learned distribution~\cite{IKD,Patchcore}. A widely adopted strategy involves utilizing pre-trained neural networks~\cite{RD4AD,2023CaiDis,CDO} to extract meaningful feature representations, which are then processed using methods such as feature reconstruction~\cite{DSR,diad,bionda2022deep,yao_feature_2022}, knowledge distillation~\cite{2023liuST,jiang_masked_2023,vitad}, or memory-based frameworks~\cite{GCPF,regad}. These approaches are particularly effective when only normal samples are available, but their performance can degrade in scenarios with complex distributions or when normal samples are insufficiently representative.

Semi-supervised AD methods~\cite{BiaS,ding2022catching} address this limitation by incorporating both normal and labeled anomalous samples during training. These methods leverage the labeled anomalies to construct more precise decision boundaries, enabling improved detection of outliers. However, this performance gain comes at the cost of requiring annotated abnormal data, which can be challenging to obtain in many practical applications. While semi-supervised methods often outperform their unsupervised counterparts under controlled conditions, their dependency on labeled data limits their scalability and applicability in diverse, real-world contexts.

Despite the progress made by traditional AD methods, their reliance on either large numbers of normal samples or annotated anomalies creates a bottleneck in data-scarce environments. To address these challenges, we leverage the rich information inherent in unlabeled data for anomaly detection. Our work establishes a zero-shot anomaly detection (ZSAD) framework that operates directly on unlabeled test set data, enabling anomaly detection in completely unseen categories without requiring additional training procedures. This approach represents a significant advancement toward more generalizable and data-efficient anomaly detection solutions.

\subsection{Zero-shot Anomaly Detection}
Zero-shot anomaly detection (ZSAD) has gained increasing attention as a solution for identifying anomalies in unseen categories without requiring task-specific training data. Leveraging the generalization capabilities of zero-shot models, many approaches have sought to extend pre-trained Vision-Language Models (VLMs) to anomaly detection tasks. These models, trained on large-scale datasets of image-text pairs, can link visual inputs to semantic concepts, making them promising tools for anomaly detection. For instance, certain methods exploit the representational power of CLIP~\cite{radford2021learning} by matching image features with textual descriptions of normal and anomalous states, while others refine this process through strategies such as text augmentation or region localization.

However, as VLMs are not explicitly trained for anomaly detection, additional adaptations are often required. To address this, works like APRIL-GAN~\cite{APRIL-GAN} and CLIP-AD~\cite{clipad} integrate auxiliary datasets to fine-tune components of the model, such as projection layers, for better anomaly detection performance. Prompt engineering has also emerged as a significant area of innovation. Techniques like AnomalyCLIP~\cite{zhou2023anomalyclip} and AdaCLIP~\cite{cao2025adaclip} incorporate learnable or optimized text prompts to guide pre-trained models toward detecting anomalous patterns more effectively, with some focusing on supervised or semi-supervised tasks to enhance generalization.

Building upon these advances, recent research emphasizes hybrid approaches that combine visual and language prompts with auxiliary data to bridge the gap between general-purpose VLMs and specialized anomaly detection requirements. In this work, we further extend these efforts by introducing a framework with multi-level and multi-modal fusion to maximize adaptability and achieve robust ZSAD performance across diverse scenarios.

\section{Method}

\subsection{Overview}
Our method operates on unlabeled test images using only category information through three key components: First, we employ multi-scale feature aggregation ($r=1,3,5$) to extract and evaluate ViT features against text embeddings, identifying representative normal patterns (\ref{3.2}). Second, we introduce a Pseudo-Anomaly-Aware Memory module that constructs dual memory banks to capture and highlight pseudo-anomalous regions (\ref{3.3}). Finally, a Pseudo-Anomaly Decision module suppresses pseudo-anomalies through differential analysis between the memory banks, enhancing pixel-level anomaly detection accuracy (\ref{3.4}).

\subsection{Multi-Scale Feature Aggregation}\label{3.2}
Leveraging CLIP's patch-level similarity computation mechanism, we propose a multi-scale feature aggregation approach for industrial anomaly detection. For unlabeled test images, we first obtain text features $f_{\text{pos}}$ and $f_{\text{neg}}$ through CLIP's text encoder using complementary prompts (``a photo of normal/defective \{category\}''). To address the localized nature of industrial defects, we extract both class tokens ($\text{cls\_token}$) and patch-level features ($f_{\text{img}}$) through CLIP's image encoder and propose a multi-scale feature aggregation strategy using three receptive field scales.

\begin{equation}
F_r = \frac{1}{r^2}\sum_{i,j \in \mathcal{N}_r} f_{i,j}, \quad r \in \{1,3,5\}
\end{equation}

where $\mathcal{N}_r(p)$ denotes the $r \times r$ neighborhood centered at position $p$. The similarity scores between aggregated features and normal text embedding are computed as:

\begin{equation}
S_r = \frac{F_r \cdot f_{\text{pos}}}{||F_r|| \cdot ||f_{\text{pos}}|| \cdot \tau}
\end{equation}

where $\tau$ is a temperature parameter. We further select the most normal images (top~10\%) based on global similarity:

\begin{equation}
S_{\text{cls}} = \frac{\text{sim}(\text{cls\_token}, f_{\text{pos}})}{\tau}
\end{equation}

This multi-scale approach enables effective detection of both small defects ($r=1$) and broader contextual patterns ($r=3,5$), achieving robust discrimination between true defects and environmental variations.
\subsection{Pseudo-Anomaly-Aware Memory (PAM)}\label{3.3}
In this subsection, we present a Pseudo-Anomaly-Aware Memory module that constructs dual multi-scale memory banks for precise anomaly segmentation. This architecture facilitates the discrimination between genuine defects and environmental variations. Additionally, we propose a Content-Adaptive Data Augmentation module, which enhances the feature diversity and robustness of the model. The module consists of the following key components:

\subsubsection{Content-Adaptive Data Augmentation}
Data augmentation of normal samples is essential for robust anomaly detection, where optimal strategies vary across different object and texture types~\cite{lee2024text}. To address this diversity, we propose a Content-Adaptive Data Augmentation (CADA) method that quantifies object position variations through spatial distribution variance:

\begin{equation}
\sigma = \sqrt{\sigma_{\text{spatial}}^2 + \sigma_{\text{temporal}}^2}
\end{equation}

where $\sigma_{\text{spatial}}$ and $\sigma_{\text{temporal}}$ represent the standard deviations in spatial and temporal dimensions. Based on this metric, the method adaptively selects augmentation strategies: conservative strategies focusing on appearance changes for scenes with significant position variations, and more aggressive strategies with geometric transformations for fixed-position scenes. This content-adaptive mechanism ensures both diversity and authenticity of augmented data, thereby providing more reliable training samples for anomaly detection.

\subsubsection{Normal Memory Bank Construction}
Features from the most representative normal images (top~10\% based on text similarity scores) undergo Content-Adaptive Data Augmentation and extraction, followed by integration with multi-level patch features based on their spatial positions and text similarity scores. An approximate greedy algorithm is used for core set sampling to store the normal memory bank. For each layer $l$ in each aggregation level, the feature tensors ${Z_l \in \mathbb{R}^{N \times L \times C}}$ are merged, where $N$ is the number of samples, $L$ is the feature sequence length, and $C$ is the feature dimension. The features are then reshaped into patch-level representations $(N \times L, C)$ and reduced in dimension using a feature sampler. The sampled features are stored in the memory bank.

The efficacy of the normal memory bank in pseudo-anomaly detection is supported by the following theoretical foundations:
\begin{itemize}
\item Through the cosine distance calculation formula $d(f_q, f_r) = 1 - \frac{f_q \cdot f_r}{||f_q|| \cdot ||f_r||}$, the distance between normal sample features $f_r$ in the normal memory bank and query features $f_q$ reflects their content differences.
\item Query features containing pseudo-anomalies exhibit increased distance values due to significant semantic differences from the normal features in the normal memory bank, thus highlighting pseudo-anomalous regions.
\item The normal memory bank provides a reliable reference baseline for pseudo-anomaly detection by retaining the feature representations of the most normal samples.
\end{itemize}

\subsubsection{Full Memory Bank Construction}
The full memory bank construction leverages the multi-level features extracted and aggregated through the CLIP image encoder as described in Section~\ref{3.2}. These features undergo core set sampling for efficient storage while maintaining representative coverage of the feature space.

\subsection{Pseudo-Anomaly-Aware Decision (PAD)}\label{3.4}
To address the challenge of discriminating between genuine defects and environmental variations, we propose a dual memory bank difference strategy. This strategy decouples the representation of background changes and real anomalies by designing a normal memory bank ($M_{\text{Normal}}$) and a full memory bank ($M_{\text{Full}}$). The core idea of this decoupling design is that the normal memory bank primarily captures background variation patterns (e.g., lighting, viewpoint changes) in normal samples, while the full memory bank includes both background information and potential defect features.

\subsubsection{Feature Matching and Score Computation}
For a given query sample $X$, we first transform it into a feature representation $Z_q \in \mathbb{R}^{1 \times L \times C}$. The inference process comprises several sequential steps:

1. Distance Computation:
\begin{itemize}
\item Compute the cosine distance between each query patch and all reference features in both memory banks
\item Determine the initial anomaly score by selecting the minimum distance for each query patch
\item Apply adaptive interval averaging strategy, where the final anomaly score is derived from the larger value among the top-$k$ minimum distances, with $k$ bounded by $[k_{\text{min}}, k_{\text{max}}]$
\end{itemize}

2. Anomaly Response Generation:
\begin{equation}
\begin{split}
S_c(x) &= \Psi(x, M_c), \quad x \in X \\
&\text{(Full Anomaly Response)}
\end{split}
\end{equation}

where the anomaly scoring function $\Psi(\cdot)$ is based on $K$-nearest neighbor distance:
\begin{equation}
\begin{split}
\Psi(x, M) = \text{mean}(&\text{topk}(\text{min}(D(x, m)), k)) \\
&\text{for} \quad m \in M 
\end{split}
\end{equation}

The sample-level anomaly score is ultimately obtained by averaging the patch-level scores.

\subsubsection{Content Adaptive Subtraction}
To fully utilize the complementarity of the dual memory bank, we design an adaptive differential fusion strategy. The core expression is:

\begin{equation}
\begin{split}
S_{\text{final}}(x) = S_c(x) - \alpha \cdot S_b(x) \cdot I(S_b(x) \leq \tau)
\end{split}
\end{equation}

This strategy consists of three key components:
\begin{itemize}
\item \textbf{Adaptive Threshold} ($\tau$): Dynamically determined based on the background response distribution to ensure the adaptability of the differential operation.
\item \textbf{Differential Coefficient} ($\alpha$): Ranges from $[0,1]$, controlling the intensity of background suppression.
\item \textbf{Indicator Function} ($I(\cdot)$): Implements selective differential operation to prevent over-suppression of real defect features.
\end{itemize}

The differential strategy exhibits the following characteristics:
\begin{itemize}
\item \textbf{Adaptability}: The differential intensity automatically adjusts based on the strength of the background anomaly response, ensuring the appropriate suppression of varying levels of background noise.
\item \textbf{Protection Mechanism}: The threshold control mechanism ensures that the real defect responses are not excessively suppressed.
\item \textbf{Locality}: The differential operation is performed at the pixel level, maintaining the fineness of the detection results.
\end{itemize}

\subsubsection{Multi-Scale Fusion}
To further enhance the stability and robustness of the detection, we perform differential detection at multiple feature scales ($R$):

\begin{equation}
S = \sum_{r \in R} (w_r \cdot S_r)
\end{equation}

where $w_r$ is the weight coefficient for each scale. multi-scale fusion allows for the comprehensive utilization of feature information at different scales, improving detection accuracy.

Empirical evaluations demonstrate that the proposed dual memory bank fusion strategy achieves superior performance in challenging scenarios with complex backgrounds and variations in illumination. Particularly in the presence of complex background textures and lighting variations, the adaptive differential mechanism significantly reduces false-positive rates while maintaining high defect detection rates. This performance improvement is primarily attributed to the decoupling representation capability of the dual memory bank and the effectiveness of the adaptive differential strategy.

\begin{table*}[t]
\small
\setlength{\tabcolsep}{10pt}  
\centering
\label{tab:results}
\begin{tabular}{ccccccccc}
\hline
Methods & Setting & AUROC-cls & F1-max-cls & AP-cls & AUROC-segm & F1-max-segm & AP-segm & PRO-segm \\
\hline
WinCLIP & 0-shot & 91.8 & 92.9 & 96.5 & 85.1 & 31.7 & - & 64.6 \\
AnomalyCLIP & 0-shot & 91.5 & - & 96.2 & 91.1 & - & - & 81.4 \\
PromptAD & 0-shot & 90.8 & - & - & 92.1 & 36.2 & - & 72.8 \\
AdaCLIP & 0-shot & 89.2 & 90.6 & - & 88.7 & 43.4 & - & - \\
VCP-CLIP & 0-shot & - & - & - & 92.0 & - & 49.4 & 87.3 \\
MuSc & 0-shot & 97.8 & 97.4 & 99.1 & 97.1 & 62.2 & 62.3 & 93.5 \\
WinCLIP & 4-shot & 95.2 & 94.7 & 97.3 & 96.2 & 51.7 & - & 88 \\
MVFA-AD & 4-shot & 96.2 & - & - & 96.3 & - & - & - \\
PromptAD & 4-shot & 96.6 & - & - & 96.5 & - & - & - \\
RegAD & 4-shot & 89.1 & 92.4 & 94.9 & 96.2 & 51.7 & 48.3 & 88 \\
\hline
Ours & 0-shot & \textbf{98.4} & \textbf{97.8} & \textbf{99.3} & \textbf{97.5} & \textbf{63.0} & \textbf{63.4} & \textbf{93.7} \\
\hline
\end{tabular}
    \makebox[\textwidth][c]{%
        \begin{minipage}{\textwidth}
            \centering
            \caption{Quantitative results on the MVTec AD dataset (\%).}
        \end{minipage}
    }
\end{table*}

\begin{table*}[t]
\small
\setlength{\tabcolsep}{10pt}  
\centering
\label{tab:results}
\begin{tabular}{ccccccccc}
\hline
Methods & Setting & AUROC-cls & F1-max-cls & AP-cls & AUROC-segm & F1-max-segm & AP-segm & PRO-segm \\
\hline
WinCLIP & 0-shot & 78.1 & 79 & 81.2 & 79.6 & 14.8 & - & 56.8 \\
AnomalyCLIP & 0-shot & 82.1 & - & 85.4 & 95.5 & - & - & 87 \\
AdaCLIP & 0-shot & 85.8 & 83.1 & - & 95.5 & 37.7 & - & - \\
VCP-CLIP & 0-shot & - & - & - & 95.7 & - & 30.1 & 90.7 \\
MuSc & 0-shot & 92.6 & 89.1 & 93.3 & 98.7 & 48.9 & 45.4 & \textbf{92.4} \\
PromptAD & 4-shot & 89.1 & - & - & 97.4 & - & - & - \\
WinCLIP & 4-shot & 87.3 & 84.2 & 88.8 & 97.2 & 47 & - & 87.6 \\
\hline
Ours & 0-shot & \textbf{93.4} & \textbf{89.7} & \textbf{93.9} & \textbf{98.8} & \textbf{50.3} & \textbf{46.9} & \textbf{92.4} \\
\hline
\end{tabular}
    \makebox[\textwidth][c]{%
        \begin{minipage}{\textwidth}
            \centering
            \caption{Quantitative results on the VisA dataset (\%).}
        \end{minipage}
    }
\end{table*}

\begin{figure*}[t]
    \centering
    \includegraphics[width=1.0\linewidth]{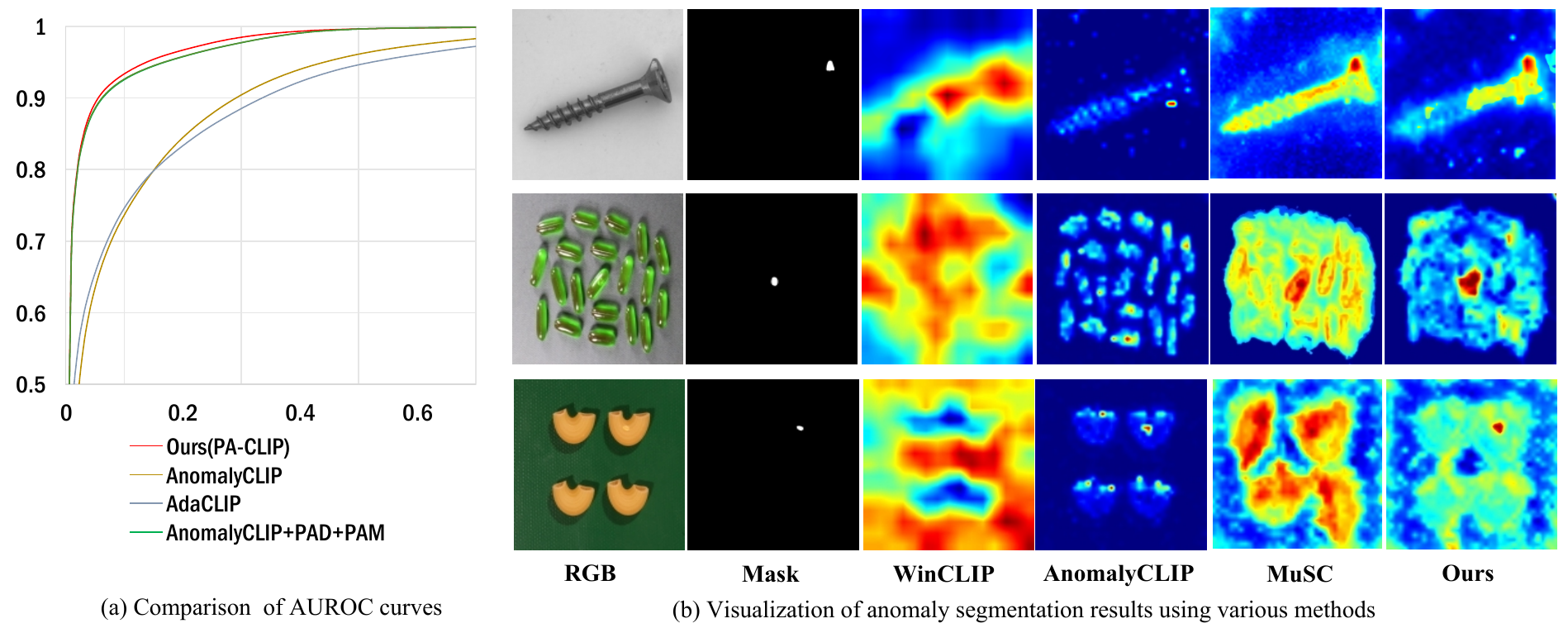}
    \caption{
        Performance comparison of anomaly detection methods, including AUROC metrics and segmentation visualization.}
    \label{'vis'}
        \vspace{-1em}
\end{figure*}

\section{Experiments}
\subsection{Experimental Setting}
All experiments are conducted on H100,We employ VIT-L-14-336 as the backbone architecture with input images resized to 518×518. This model comprises 24 feature layers, and we select layers 5, 11, 17, and 23, which correspond to the linear projection layers of VIT, to perform our tasks.

\subsubsection{Datasets} 
We evaluate our method on two widely-used industrial anomaly detection benchmarks:
\begin{itemize}
\item \textbf{MVTec AD} \cite{MVTec-AD}: Contains high-resolution RGB images (700×700 to 1024×1024) across 10 object categories and 5 texture categories.
\item \textbf{VisA} \cite{zou2022spot}: Comprises high-resolution RGB images (1000×1500) of 12 objects spanning 3 different domains.
\end{itemize}
Both datasets provide comprehensive evaluation protocols with normal training samples and anomalous test cases.

\subsubsection{Evaluation Metrics} 
For classification metrics, we report three metrics:
Area Under Receiver Operating Characteristic curve (AUROC)
Average Precision (AP)
F1-score at the optimal threshold (F1-max)

For segmentation metrics, we report four metrics:
Pixel-wise AUROC
Pixel-wise F1-max
Pixel-wise AP
Per-Region Overlap (PRO)

\subsubsection{Baselines}
We compare our method against state-of-the-art zero-shot/few-shot approaches:
\begin{itemize}
\item CLIP-based methods: WinCLIP \cite{WinClip}, AnomalyCLIP \cite{zhou2023anomalyclip}, PromptAD \cite{li2024promptad}, AdaCLIP \cite{cao2025adaclip}, VCP-CLIP \cite{qu2024vcp}
\item Other approaches: MuS c \cite{li2024musc}, MVFA-AD \cite{MVTec-AD}, RegAD \cite{regad}, RegAD+Adversarial Loss \cite{lee2023few}
\end{itemize}
All reported results are obtained from the official implementations of respective methods.
\subsection{Experimental Results} 

Experimental results in Tables 1 and 2 highlight the superiority of our method compared to existing zero-shot and few-shot approaches across both MVTec AD and VisA benchmarks. On MVTec AD, our method achieves state-of-the-art performance in classification with a 98.4\% AUROC, surpassing zero-shot methods such as MuSc (97.8\%), WinCLIP (91.8\%), and AnomalyCLIP (91.5\%). In segmentation tasks, we achieve 97.5\% AUROC and 93.7\% PRO score, outperforming even few-shot methods like WinCLIP (4-shot) and PromptAD (4-shot).

On the challenging VisA dataset, we achieve 93.4\% classification AUROC and 98.8\% segmentation AUROC, significantly outperforming WinCLIP in both classification (93.4\% vs. 78.1\%) and segmentation (98.8\% vs. 79.6\%), while also showing competitive performance against MuSc in segmentation PRO (92.4\%). Figure 3 demonstrates our method's robust anomaly localization capabilities across various object categories.

\subsection{Ablation Study}
To systematically evaluate the contribution of each architectural component, we conduct extensive ablation studies on both benchmark datasets. Starting from a baseline implementation, we progressively incorporate our proposed modules to quantify their individual and combined effects. This step-by-step analysis provides a comprehensive understanding of each component's role and its synergy within the overall architecture.

\subsubsection{Effectiveness of Key Components}
As shown in Tables 3 and 4, our analysis reveals the following contributions:

\begin{table}[H]
\small
\setlength{\tabcolsep}{2pt}  
\centering
\label{tab:results}
\begin{tabular}{ccccc}
\hline
Methods  & AUROC-cls & AP-cls & AUROC-segm & PRO-segm \\
\hline
baseline  & 91.8 & 96.5 & 85.1 & 64.6 \\
+PAD  & 97.4 & 98.9 & 97.3 & 93.4 \\
+PAD+PAM  & 98.1 & 99.2 & 97.3 & 93.4 \\
+PAD+MPAM  & 98.4 & 99.3 & 97.5 & 93.7 \\
\hline
\end{tabular}
\caption{Ablation results on the mvtec dataset (\%).}
\end{table}

\begin{table}[H]
\small
\setlength{\tabcolsep}{2pt}  
\centering
\label{tab:results}
\begin{tabular}{ccccc}
\hline
Methods  & AUROC-cls & AP-cls & AUROC-segm & PRO-segm \\
\hline
baseline   & 78.1 & 81.2 & 79.6 & 56.8 \\
+PAD  & 92.3 & 93.2 & 98.7 & 92.4 \\
+PAD+PAM  & 92.9 & 93.5 & 98.4 & 92.2 \\
+PAD+MPAM & 93.4 & 93.9 & 98.8 & 92.4 \\
\hline
\end{tabular}
\caption{Ablation results on the visa dataset (\%).}
 \vspace{-1em}

\end{table}

\noindent\textbf{Impact of Pseudo-Anomaly Decision Module} 
Starting from the baseline, adding our Pseudo-Anomaly-Aware Decision Module (PAD) module brings substantial improvements. On MVTec AD, it improves the classification AUROC from 91.8\% to 97.4\% (+5.6\%) and segmentation AUROC from 85.1\% to 97.3\% (+12.2\%). Similar significant gains are observed on VisA, with classification AUROC increasing from 78.1\% to 92.3\% (+14.2\%) and segmentation AUROC from 79.6\% to 98.7\% (+19.1\%). These results demonstrate that PA effectively enhances the model's ability to distinguish normal and anomalous patterns.

\noindent\textbf{Enhancement with Pseudo-Anomaly-Aware Memory} 
Building upon PAD, the addition of Pseudo-Anomaly-Aware Memory (PAM) further improves the model's performance, particularly in classification tasks. On MVTec AD, it increases classification AUROC to 98.1\% (+0.7\%) while maintaining strong segmentation performance. The improvements are also consistent on VisA, with classification AUROC reaching 92.9\% (+0.6\%).

\noindent\textbf{Benefits of Multi-Scale PAM.} 
The final incorporation of Multi-Scale PAM (MPAM) leads to our best performing model. This module brings additional improvements across all metrics, achieving 98.4\% classification AUROC and 97.5\% segmentation AUROC on MVTec AD, and 93.4\% classification AUROC and 98.8\% segmentation AUROC on VisA. The multi-scale design helps capture anomalies at different granularities, resulting in more robust detection performance.

\subsubsection{Generalization to Other CLIP-based Methods}
To verify the generalization ability of our proposed modules, we apply our PAD and PAM components to other CLIP-based methods (AdaCLIP and AnomalyCLIP). As shown in Tables 5 and 6, our modules consistently improve the performance of these methods:

\begin{table}[H]
\small
\setlength{\tabcolsep}{2pt}  
\centering
\label{tab:results}
\begin{tabular}{ccccc}
\hline
Methods  & AUROC-cls & AP-cls & AUROC-segm & PRO-segm \\
\hline
AdaCLIP  & 91.8 & 96.5 & 85.1 & 64.6 \\
+PAD+PAM  & 97.2 & 98.8 & 96.9 & 92.8 \\
AnomalyCLIP  & 91.5 & 96.2 & 91.1 & 81.4 \\
+PAD+PAM  & 97.3 & 98.9 & 96.9 & 92.8 \\
\hline
\end{tabular}
\caption{CLIP-based methods on the MVTec AD dataset (\%).}
        \vspace{-1em}

\end{table}

\begin{table}[H]
\small
\setlength{\tabcolsep}{2pt}  
\centering
\label{tab:results}
\begin{tabular}{ccccc}
\hline
Methods  & AUROC-cls & AP-cls & AUROC-segm & PRO-segm \\
\hline
AdaCLIP  & 85.8 & - & 95.5 & - \\
+PAD+PAM  & 92.9 & 93.7 & 98.5 & 91.3 \\
AnomalyCLIP  & 82.1 & 85.4 & 95.5 & 87 \\
+PAD+PAM  & 92.5 & 93.3 & 98.7 & 92.4 \\
\hline
\end{tabular}
\caption{CLIP-based methods on the VisA dataset (\%).}
        \vspace{-1em}

\end{table}

\section{Summary}
In this paper, we introduce PA-CLIP, a novel zero-shot anomaly detection method that effectively integrates CLIP to combine textual and visual information. Our approach delves into the representation of background noise through the lens of Pseudo-Anomaly awareness. Initially, we employ Multiple Aggregation CLIP to identify and extract the most representative normal patches and full patches, guided by this awareness. Subsequently, we develop Pseudo-Anomaly-Aware Memory to enhance feature representation. Finally, we implement a Pseudo-Anomaly-Aware Decision method to further refine detection outcomes. Importantly, our method not only adeptly mitigates the influence of pseudo-anomalies but also outperforms existing zero-shot anomaly detection techniques, as well as many few-shot methods. This advancement underscores the potential of PA-CLIP in achieving robust and accurate anomaly detection in complex industrial environments.
\clearpage
\appendix

\bibliographystyle{named}  
\bibliography{reference}  

\end{document}